\documentclass[conference]{IEEEtran}
\IEEEoverridecommandlockouts
\usepackage{cite}
\usepackage{amsmath,amssymb,amsfonts}
\usepackage{algorithmic}
\usepackage{graphicx}
\usepackage{textcomp}
\usepackage{xcolor}
\usepackage{booktabs}
\usepackage{multirow}
\usepackage{kotex}
\usepackage{makecell}
\usepackage{graphicx, wrapfig}
\usepackage{capt-of}
\def\BibTeX{{\rm B\kern-.05em{\sc i\kern-.025em b}\kern-.08em
    T\kern-.1667em\lower.7ex\hbox{E}\kern-.125emX}}

\newcommand{\figref}[1]{Fig.~\ref{#1}}
\newcommand{\tabref}[1]{Table~\ref{#1}}

\makeatletter
\newcommand{\linebreakand}{%
  \end{@IEEEauthorhalign}
  \hfill\mbox{}\par
  \mbox{}\hfill\begin{@IEEEauthorhalign}
}
\makeatother
\begin{document}

\title{Bridging Geometric and Semantic Foundation Models for Generalized Monocular Depth Estimation
\thanks{\textsuperscript{$\ast$}Equal Contribution}
\thanks{\textsuperscript{$\dagger$}Corresponding Author}
}

\author{\IEEEauthorblockN{Sanggyun Ma\textsuperscript{$\ast$}}
\IEEEauthorblockA{\textit{DGIST}\\
Daegu, South Korea \\
wodon-326@dgist.ac.kr}
\and
\IEEEauthorblockN{Wonjoon Choi\textsuperscript{$\ast$}}
\IEEEauthorblockA{\textit{DGIST}\\
Daegu, South Korea \\
wjchoi@dgist.ac.kr}
\and
\IEEEauthorblockN{Jihun Park\textsuperscript{$\ast$}}
\IEEEauthorblockA{\textit{DGIST}\\
Daegu, South Korea \\
pjh2857@dgist.ac.kr}
\and
\IEEEauthorblockN{Jaeyeul Kim}
\IEEEauthorblockA{\textit{DGIST}\\
Daegu, South Korea \\
jykim94@dgist.ac.kr}
\linebreakand 
\IEEEauthorblockN{Seunghun Lee}
\IEEEauthorblockA{\textit{DGIST}\\
Daegu, South Korea \\
lsh5688@dgist.ac.kr}
\and
\IEEEauthorblockN{Jiwan Seo}
\IEEEauthorblockA{\textit{DGIST}\\
Daegu, South Korea \\
eccaron@dgist.ac.kr}
\and
\IEEEauthorblockN{Sunghoon Im\textsuperscript{$\dagger$}}
\IEEEauthorblockA{\textit{DGIST}\\
Daegu, South Korea \\
sunghoonim@dgist.ac.kr}
}
\maketitle

\begin{abstract}
We present Bridging Geometric and Semantic (BriGeS), an effective method that fuses geometric and semantic information within foundation models to enhance Monocular Depth Estimation (MDE).
Central to BriGeS is the Bridging Gate, which integrates the complementary strengths of depth and segmentation foundation models.
This integration is further refined by our Attention Temperature Scaling technique.
It finely adjusts the focus of the attention mechanisms to prevent over-concentration on specific features, thus ensuring balanced performance across diverse inputs.
BriGeS capitalizes on pre-trained foundation models and adopts a strategy that focuses on training only the Bridging Gate. 
This method significantly reduces resource demands and training time while maintaining the model's ability to generalize effectively.
Extensive experiments across multiple challenging datasets demonstrate that BriGeS outperforms state-of-the-art methods in MDE for complex scenes, effectively handling intricate structures and overlapping objects.
\end{abstract}

\begin{IEEEkeywords}
Foundation model, Monocular depth estimation, Zero-shot depth estimation
\end{IEEEkeywords}
\section{Introduction}
\label{sec:intro}

\begin{figure}[t]
    \centering
    \includegraphics[width=0.9\linewidth]{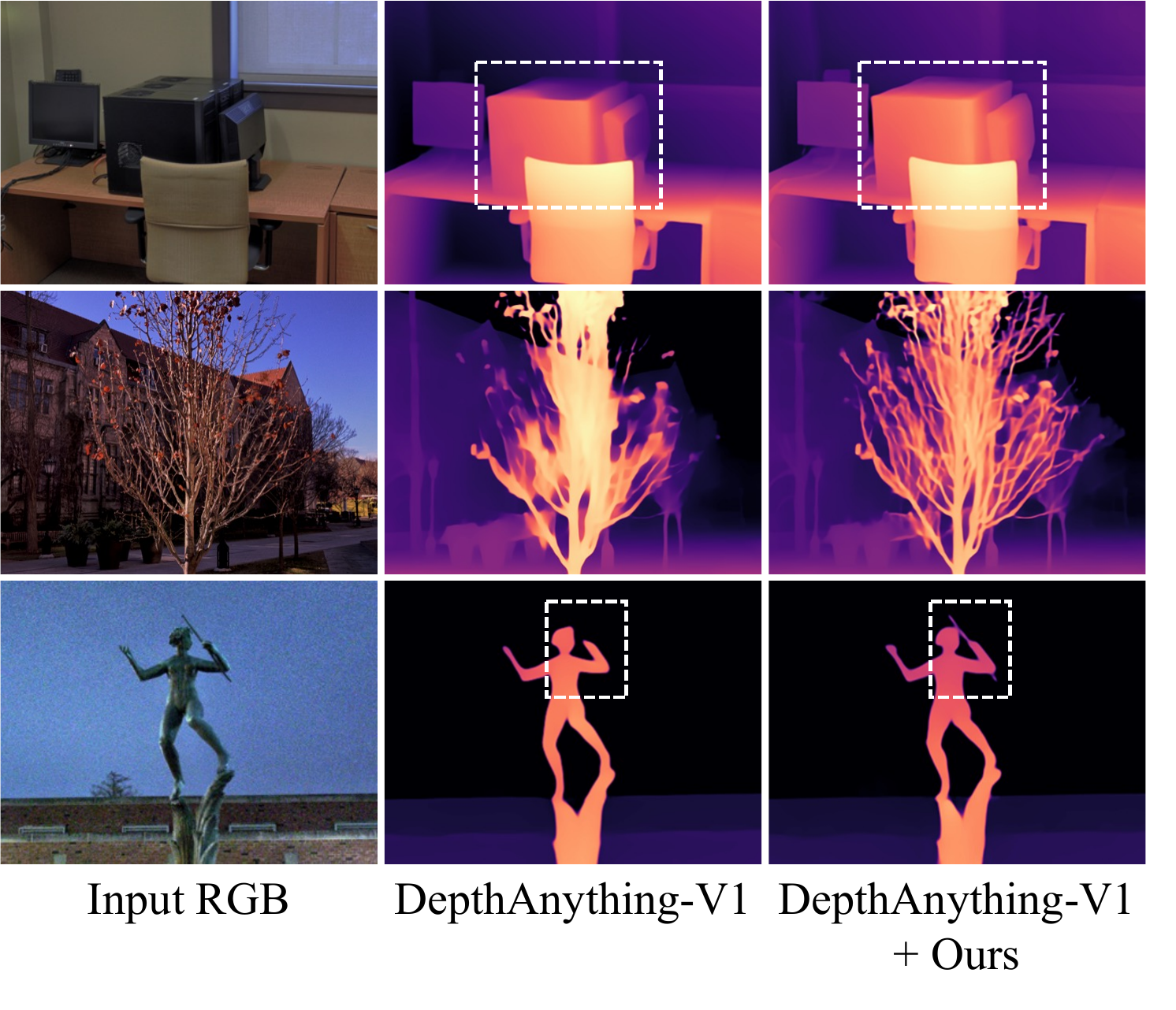}
    \vspace{-0.6em}
    \caption{Qualitative comparison with DepthAnything-V1 \cite{yang2024depth}.}
    \label{fig:compare_figure_v1}
    \vspace{-1.2em}
\end{figure}

Monocular Depth Estimation (MDE) is a crucial task that predicts the depth map from a single image. This method is cost-effective and simple to deploy.
Utilizing just one camera eliminates the need for complex hardware setups or additional camera installations, thereby reducing initial expenses and simplifying maintenance.
Given its simple structure, it can be seamlessly integrated across various environments and platforms.
This adaptability makes it especially valuable in fields like robotics, autonomous driving, and augmented reality.

Historically, MDE relies on handcrafted features and traditional vision techniques \cite{hoiem2007recovering}.
However, the advent of CNN \cite{eigen2014depth} and ViT-based models \cite{ranftl2021vision} has significantly enhanced its performance. 
Recent studies \cite{klingner2020self} have highlighted the advantages of incorporating semantic context from images, which provide more informative cues than solely relying on local geometry.
More recently, MDE works \cite{yang2024depth, yang2024depthv2, ke2024repurposing} have focused on training large-scale models using a vast amount of training datasets inspired by foundation models \cite{bommasani2021opportunities, ranftl2020towards, kirillov2023segment}.
These approaches aim to harness the broader contextual understanding inherent in large pre-trained models, significantly advancing the field of depth estimation.

Despite these advancements, current foundation models in MDE \cite{yang2024depth, yang2024depthv2, ke2024repurposing} have primarily utilized geometric data for depth estimation and have not directly incorporated semantic information.
These models require substantial training data and considerable computational resources, making it challenging to integrate additional semantic information.
In response to these challenges, we propose a new integration module for MDE that adaptively incorporates semantic information, named Bridging Geometric and Semantic (BriGeS).
This approach is particularly effective in addressing complex visual elements such as intricate structures, homogeneous regions, and delicate structures as shown in \figref{fig:compare_figure_v1}.

To integrate depth features with semantic features under a resource-constrained training environment, we design Bridging Gate, which directly fuses features from the depth \cite{yang2024depth, yang2024depthv2} and segmentation \cite{kirillov2023segment} foundation model.
This gate employs cross-attention and self-attention blocks to facilitate effective feature integration.
To ensure high-depth estimation performance while preserving prior knowledge of a model trained on large-scale data, we exclusively train the Bridging Gate, keeping all encoders and decoders frozen.
Additionally, we introduce the Attention Temperature Scaling technique to adjust the variance of the Bridging Gate's attention map.
This adjustment is crucial as the map tends to become overly concentrated on specific regions due to the fusion of two distinct modalities.
By altering the distribution of attention operations during the inference stage, the technique mitigates this issue, thereby simplifying the process and enhancing overall effectiveness.

Extensive experiments demonstrate significant performance improvements achieved by employing BriGeS on unseen datasets, particularly excelling in images with complex structures.
Our contributions to the field can be summarized as follows:
\begin{itemize}
    \item {We present BriGeS, an effective module that fuses a depth foundation model and a segmentation foundation model using minimal data and training effort to enhance Monocular Depth Estimation.}
    \item {We propose the Bridging Gate, an adaptive fusion layer specifically tailored to integrate geometric and semantic information.}
    \item {We introduce the Attention Temperature Scaling, which regulates the attention distribution to mitigate over-concentration on specific regions during inference.}
\end{itemize}

\section{Method}

\subsection{Overall Pipeline}
\begin{figure*}[t]
    \centering
    \includegraphics[width=0.83\linewidth]{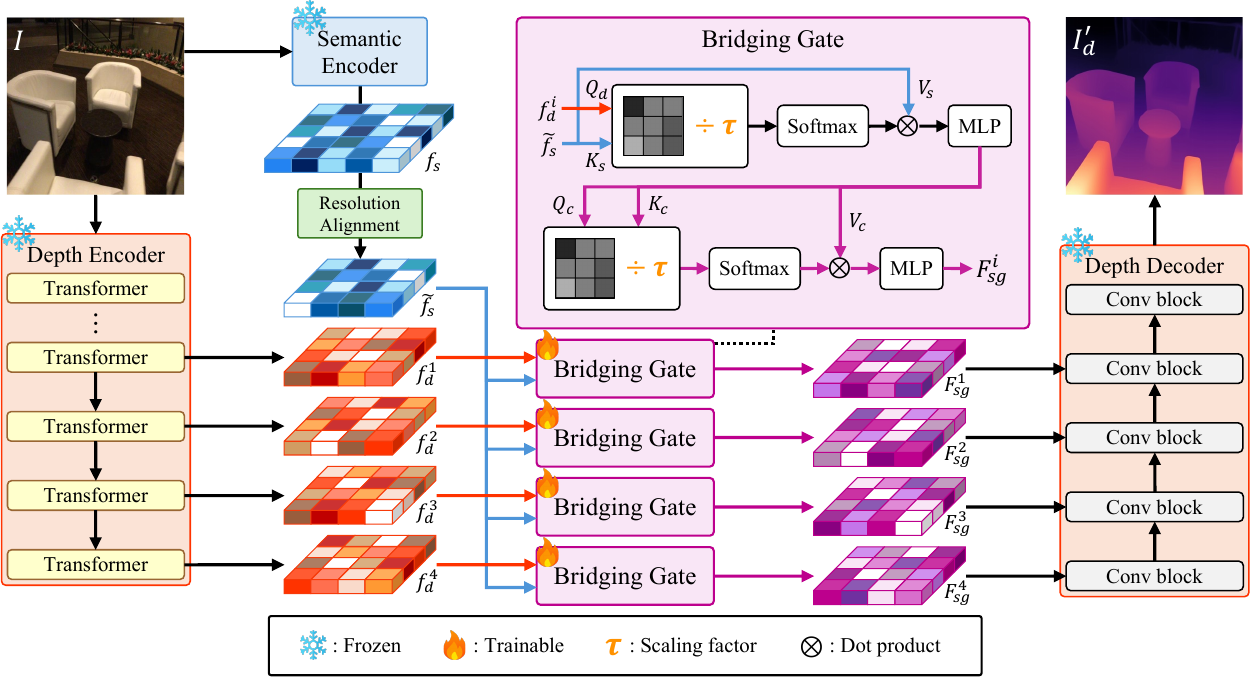}
    \vspace{-0.5em}
    \caption{Overview of the proposed BriGeS framework.}
    \label{fig:main}
    \vspace{-1.0em}
\end{figure*}

In this section, we describe the overall pipeline of BriGeS, illustrated in \figref{fig:main}.
Our BriGeS utilizes the encoders $E_d$ from DepthAnything \cite{yang2024depth, yang2024depthv2} and $E_s$ from SegmentAnything \cite{kirillov2023segment}, along with the DepthAnything decoder $D_d$.
We extract depth features $\{f^i_d\}_{i=1}^{4}$ and semantic feature $f_s$ from the respective encoders, and fuse them using Bridging Gates $\{G^i\}_{i=1}^{4}$ to generate semantic-aware geometric features $\mathbf{F}_{sg}=\{F^i_{sg}\}_{i=1}^{4}$.
Since the spatial resolutions of the semantic feature $f_s$ and the depth feature $f_d^i$ differ, we align $f_s$ to match $f_d^i$ by applying bilinear interpolation followed by max pooling, resulting in the resolution-aligned semantic feature $\Tilde{f}_s$.
The same resolution-aligned semantic feature $\Tilde{f}_s$ is shared across all Bridging Gates.
The decoder then leverages these semantic-aware geometric features to regress the depth map $I'_d$ of the input image as follows:
\begin{equation}
\begin{gathered}
    I'_d = D_d(\mathbf{F}_{sg}), ~ \mathbf{F}_{sg} = \{G^i(f^i_d, \Tilde{f}_s)\}_{i=1}^{4}, \\
    ~ I'_d \in \mathbb{R}^{H^{\prime} \times W^{\prime}}, ~ F^i_{sg} \in \mathbb{R}^{HW \times C}.
\end{gathered}
\end{equation}

From the following section, we simplify the notation by omitting the subscript $i$ (e.g., $f^i_d$ is written as $f_d$) to streamline the equations and discussion.


\subsection{Bridging Gate}
\label{section:Bridging_Gate}
While DepthAnything \cite{yang2024depth, yang2024depthv2} demonstrates superior generalization capabilities across a wide range of unseen scenes, its lack of semantic integration can result in over-smoothed predictions in scenarios involving complex structures or ambiguous boundaries.
To address this, we design the Bridging Gate that enables direct interaction between geometric and semantic information.
This enhancement aims to provide more accurate depth estimations, particularly in scenes characterized by complex structures or unclear object boundaries.
Our Bridging Gate ($G$) consists of a cross-attention block ($\operatorname{Block}_c$) and a self-attention block ($\operatorname{Block}_s$), designed to dynamically fuse geometric and semantic information as follows:
\begin{equation}
    G(f_{d},\Tilde{f}_s) = \operatorname{Block}_s\left(\operatorname{Block}_c(f_{d},\Tilde{f}_s)\right) .
\end{equation}

\noindent\textbf{Cross-Attention Block.}
In the cross-attention block $\operatorname{Block}_c$, the depth feature $f_d$ is used as the query, while the semantic feature $\Tilde{f}_s$ serves as both the key and value.
This configuration allows for the direct fusion of geometric and semantic information.
The resulting fused feature is further refined through an MLP layer to generate $F_c$ as follows:
\begin{equation}
\begin{gathered} 
    F_c = \operatorname{Block}_c\left(f_d, \Tilde{f}_s\right),\\
    \operatorname{Block}_c\left(f_d, \Tilde{f}_s\right) = \operatorname{MLP}\left(\operatorname{Softmax}\left(\frac{Q_d K_s^{T}}{\sqrt{d}}\right)\cdot V_s \right),\\
    Q_d=W_q^c(f_d),~K_s=W_k^c(\Tilde{f}_s),~ V_s=W_v^c(\Tilde{f}_s),
\end{gathered}
\end{equation}
where $d$ denotes the dimension of the projected query.
The weight matrices $W_q^c$, $W_k^c$, and $W_v^c$ are used to transform features $f_d$ and $\Tilde{f}_s$ to the query, key, and value, respectively.

\noindent\textbf{Self-Attention Block.}
Subsequently, the generated $F_c$ is further processed in the self-attention block $\operatorname{Block}_s$.
In this step, $F_c$ serves as the query, key, and value, enabling the self-attention mechanism to refine the feature.
The output is then passed through an MLP to produce the final semantic-aware geometric feature $F_{sg}$ as follows:
\begin{equation}
\begin{gathered}
    F_{sg} = \operatorname{Block_s}\left(F_c\right),\\
    \operatorname{Block_s}\left(F_c\right) = \operatorname{MLP}\left(\operatorname{Softmax}\left(\frac{Q_{c} K_{c}^{T}}{\sqrt{d}}\right)\cdot V_{c} \right),\\
    Q_{c}=W_q^s(F_c),\quad K_{c}=W_k^s(F_c),\quad V_{c}=W_v^s(F_c),
\end{gathered}
\end{equation}
where $W_q^s$, $W_k^s$, and $W_v^s$ are the weight matrices that transforms a feature $F_c$ to the query, key and value, respectively.
This feature represents a more precise fusion of geometric and semantic information.

\subsection{Attention Temperature Scaling}
\label{subsection:attention_temperature_scaling}
The Bridging Gate effectively merges features from two different features ($f_d$, $\Tilde{f}_s$) through attention mechanisms.
However, it can neglect peripheral details within the $f_d$ feature, as attention may become overly concentrated on central object regions.
To address this issue, we leverage the threshold \(\tau\) within the Bridging Gate's attention operations during the inference phase as follows:

\begin{equation}
\begin{gathered}
    \operatorname{Attn}_\tau\left(Q, K, V\right)=\operatorname{softmax}\left(\frac{{Q K^T}}{\tau \sqrt{d}}\right)\cdot V,~\tau>1,
\end{gathered}
\end{equation}
where $Q$, $K$, and $V$ represent query, key, and value from either the cross-attention block (\(Q_d, K_s, V_s\)) or the self-attention block (\(Q_c, K_c, V_c\)). 
By dividing the scaled-dot product between the query and key by $\tau$, this approach effectively broadens the distribution of attention from highly concentrated areas to more peripheral regions, facilitating a more comprehensive assessment of both central and surrounding features.
This redistribution of attention, applied during the inference phase, enhances the model's ability to incorporate overall structure, thereby improving the overall depth estimation performance.

\subsection{Training}
\label{section:Training}
Building a foundation model with high generalization capabilities typically requires extensive amounts of data and substantial computational resources.
However, to improve performance with minimal data and effort, we adopt a strategy focused on optimizing key components effectively.
Specifically, we train only the Bridging Gate, keeping the encoders ($E_s$, $E_d$) and decoder ($D_d$) parameters frozen, as shown in \figref{fig:main}.
This approach is designed to generate enhanced scene understanding features by combining geometric and semantic information within the Bridging Gate.

\section{Experiments}

\begin{table*}[t]
\centering 
\renewcommand{\arraystretch}{0.95}
\renewcommand\tabcolsep{4pt} 
\caption{Quantitative comparison of zero-shot relative depth estimation. (\textbf{Bold}: best, \underline{Underlined}: second best)}
\vspace{-1.0em}
\centering
\resizebox{\textwidth}{!}{
\begin{tabular}{cccccccccc}
\toprule
\multirow{2}{*}[-0.3em]{Method}                & \multirow{2}{*}[-0.3em]{Backbone}         & \multicolumn{2}{c}{KITTI}  & \multicolumn{2}{c}{NYUv2 }  & \multicolumn{2}{c}{ETH3D}  & \multicolumn{2}{c}{DIODE}  \\
\cmidrule(lr){3-4}\cmidrule(lr){5-6}\cmidrule(lr){7-8} \cmidrule(lr){9-10}
                                       &                                            & AbsRel\textdownarrow & $\delta_1$\textuparrow & AbsRel\textdownarrow & $\delta_1$\textuparrow & AbsRel\textdownarrow & $\delta_1$\textuparrow & AbsRel\textdownarrow & $\delta_1$\textuparrow     \\ \midrule
MiDaS          & ResNeXt-101                       & 0.236                & 0.630                 & 0.111                         & 0.885                 & 0.184                & 0.752                 & 0.332                & 0.715                 \\
DPT             & ViT-Large                         & 0.100                & 0.901                 & 0.098                         & 0.903                 & 0.078                & 0.946                 & \textbf{0.182}       & 0.758                 \\
Marigold       & Stable Diffusion                  & 0.099                & 0.916                 & 0.055                         & 0.964                 & 0.065                & 0.960                 & 0.308                & \textbf{0.773}        \\
GenPercept         & Stable Diffusion                  & 0.080                & 0.934                 & 0.058                         & 0.969                 & 0.096                & 0.959                 & 0.226                & 0.741                 \\ \midrule
DepthAnything-V1-Base          & ViT-Base              & 0.080                & 0.941                 & 0.045                         & 0.979                 & 0.055                & 0.978                 & 0.287                & 0.754                 \\
\textbf{DepthAnything-V1-Base + Ours}  & ViT-Base                          & \textbf{0.074}       & 0.947                 & \underline{0.042}             & \underline{0.982}     & 0.051                & 0.980                 & 0.243                & 0.756                 \\ \midrule
DepthAnything-V1-Large         & ViT-Large             & 0.080                & 0.947                 & \underline{0.042}             & 0.981                 & 0.057                & 0.982                 & 0.260                & 0.758                 \\
\textbf{DepthAnything-V1-Large + Ours} & ViT-Large                         & 0.076                & \textbf{0.950}        & \textbf{0.040}                & \textbf{0.984}        & \underline{0.050}    & \underline{0.983}     & \underline{0.222}    & \underline{0.761}     \\ \midrule
DepthAnything-V2-Base        & ViT-Base              & 0.084                & 0.936                 & 0.046                         & 0.977                 & 0.062                & 0.980                 & 0.281                & 0.750                  \\
\textbf{DepthAnything-V2-Base + Ours}  & ViT-Base                          & 0.080                & 0.939                 & 0.046                         & 0.977                 & 0.057                & 0.981                 & 0.259                & 0.752                 \\ \midrule
DepthAnything-V2-Large       & ViT-Large             & 0.080                & 0.944                 & 0.043                         & 0.979                 & 0.053                & \underline{0.983}     & 0.265                & 0.754                 \\
\textbf{DepthAnything-V2-Large + Ours} & ViT-Large                         & \underline{0.075}    & \underline{0.948}     & 0.044                         & 0.979                 & \textbf{0.049}       & \textbf{0.985}        & 0.249                & 0.754                 \\ \bottomrule
\end{tabular}
}
\label{tab:eval}
\vspace{-0.5em}
\end{table*}

\begin{table*}[t]
\renewcommand{\arraystretch}{0.9} 
\renewcommand\tabcolsep{4pt}
\caption{Comparison of zero-shot depth estimation on DA-2K \cite{yang2024depthv2}. All metrics are accuracy (\%). (\textbf{Bold}: best, DA: DepthAnything)}
\vspace{-1.0em}
\centering
\resizebox{1\textwidth}{!}{%
\begin{tabular}{cccccccc}
\toprule
\multicolumn{8}{c}{\textbf{Relative depth estimation model}} \\ \midrule
MiDaS \cite{ranftl2020towards} & DPT \cite{ranftl2021vision} & Marigold \cite{ke2024repurposing} & GenPercept \cite{xu2024matters} & DA-V1-B \cite{yang2024depth} & DA-V1-L \cite{yang2024depth} & DA-V2-B \cite{yang2024depthv2} & DA-V2-L \cite{yang2024depthv2} \\ \midrule
83.5 & 79.1 & 86.8 & 93.2 & 89.4 & 88.5 & 97.0 & 97.1 \\ \midrule \midrule
\multicolumn{4}{c}{\textbf{Metric depth estimation model}} & \multicolumn{4}{c}{\textbf{BriGeS (Ours)}} \\ \cmidrule(lr){0-3} \cmidrule(lr){5-8}
\multicolumn{2}{c}{UniDepth \cite{piccinelli2024unidepth}} & Metric3D-V2 \cite{hu2024metric3d} & DepthPro \cite{bochkovskii2025depth} & DA-V1-B & DA-V1-L & DA-V2-B & DA-V2-L \\ \midrule
\multicolumn{2}{c}{83.7} & 89.2 & 94.7 & 91.4 & 92.1 & 97.0 & \textbf{97.6} \\ \bottomrule
\end{tabular}
}
\label{tab:eval_da2k}
\end{table*}

\subsection{Implementation Details}
Our training methodology aligns closely with the one employed in the DepthAnything \cite{yang2024depth, yang2024depthv2}.
Initially, the ground truth depth value is transformed into a disparity space and then normalized to a range of 0 to 1.
To effectively train on datasets with varying depths, we employ an affine-invariant loss, which ignores the effects of scale and shift.
When training BriGeS with DepthAnything-V1, only the affine-invariant loss is used, whereas training with DepthAnything-V2 incorporates both the affine-invariant loss and the gradient matching loss from MiDaS \cite{ranftl2020towards} at a 1:2 ratio.

The batch size is set to 16, with an initial learning rate of $5e-5$. We use the AdamW optimizer \cite{loshchilov2017decoupled} with a linear schedule.
We conduct experiments on 8 RTX TITAN GPUs using labeled images, running for 2 epochs with DepthAnything-V1 and 1 epoch with DepthAnything-V2, respectively.
Training our model for a single epoch takes 8 hours for DepthAnything-Base and 18 hours for DepthAnything-Large.
During training, all images are resized so that the shorter side is 518 while maintaining the original aspect ratio, and then randomly cropped to a size of 518$\times$518.
In addition, the temperature scaling factor $\tau$ is set to 2.5 during the inference.
Our model contains 244M parameters based on the DepthAnything-Base, and 744M based on the DepthAnything-Large, including the SegmentAnything \cite{kirillov2023segment} encoder.

\subsection{Experimental Setup}
\noindent\textbf{Training Datasets.}
For training BriGeS, we employ a carefully curated subset of the extensive datasets featured in DepthAnything \cite{yang2024depth, yang2024depthv2}.
To enhance the model's generalization capabilities, we select three diverse datasets encompassing both indoor and outdoor environments for each model.
Our training data comprises 600K samples from BlendedMVS \cite{yao2020blendedmvs}, HRWSI \cite{xian2020structure}, and TartanAir \cite{wang2020tartanair} for DepthAnything-V1, and 582K samples from VKITTI \cite{cabon2020virtual}, Hypersim \cite{roberts2021hypersim}, and TartanAir for DepthAnything-V2, which corresponds to approximately 1\% of the training datasets originally used to train the respective DepthAnything.

\noindent\textbf{Evaluation Details.}
We perform zero-shot relative depth evaluation on four datasets: KITTI \cite{geiger2012we}, NYUv2 \cite{silberman2012indoor}, ETH3D \cite{schops2017multi}, and DIODE \cite{vasiljevic2019diode}.
Since our method predicts affine-invariant inverse depth, we restrict comparison to relative depth estimation models in this evaluation to ensure a fair comparison.
Evaluation is performed by aligning the estimated depth $\mathbf{d}^*$ with the ground truth $\mathbf{d}$ by computing scale and shift values.
These values are then used to derive an aligned depth map $\tilde{\mathbf{d}}~(=\text{scale}\times \mathbf{d}^* + \text{shift})$, which is adjusted to the same unit as the ground truth.
For quantitative assessment, we use the aligned depth and the ground truth, employing metrics such as AbsRel (absolute relative error: $\frac{1}{M} \sum_{i=1}^M\left|\tilde{\mathbf{d}}_i-\mathbf{d}_i\right| / \mathbf{d}_i$) and $\delta_1$ accuracy (percentage of pixels where $\max \left(\tilde{\mathbf{d}}_i / \mathbf{d}_i, \mathbf{d}_i / \tilde{\mathbf{d}}_i\right)<1.25$).
We further evaluate our method on DA-2K \cite{yang2024depthv2}, a high-resolution benchmark comprising diverse scenes, which is well-suited for analyzing fine-grained details and generalization performance.
In this benchmark, the accuracy is measured by identifying the closer of two distinct pixels.
Given this evaluation protocol based on pairwise depth comparison, we extend our evaluation to include metric depth estimation models, enabling a comprehensive assessment of our method.

\begin{figure*}[t]
    \centering
    \includegraphics[width=0.9\linewidth]{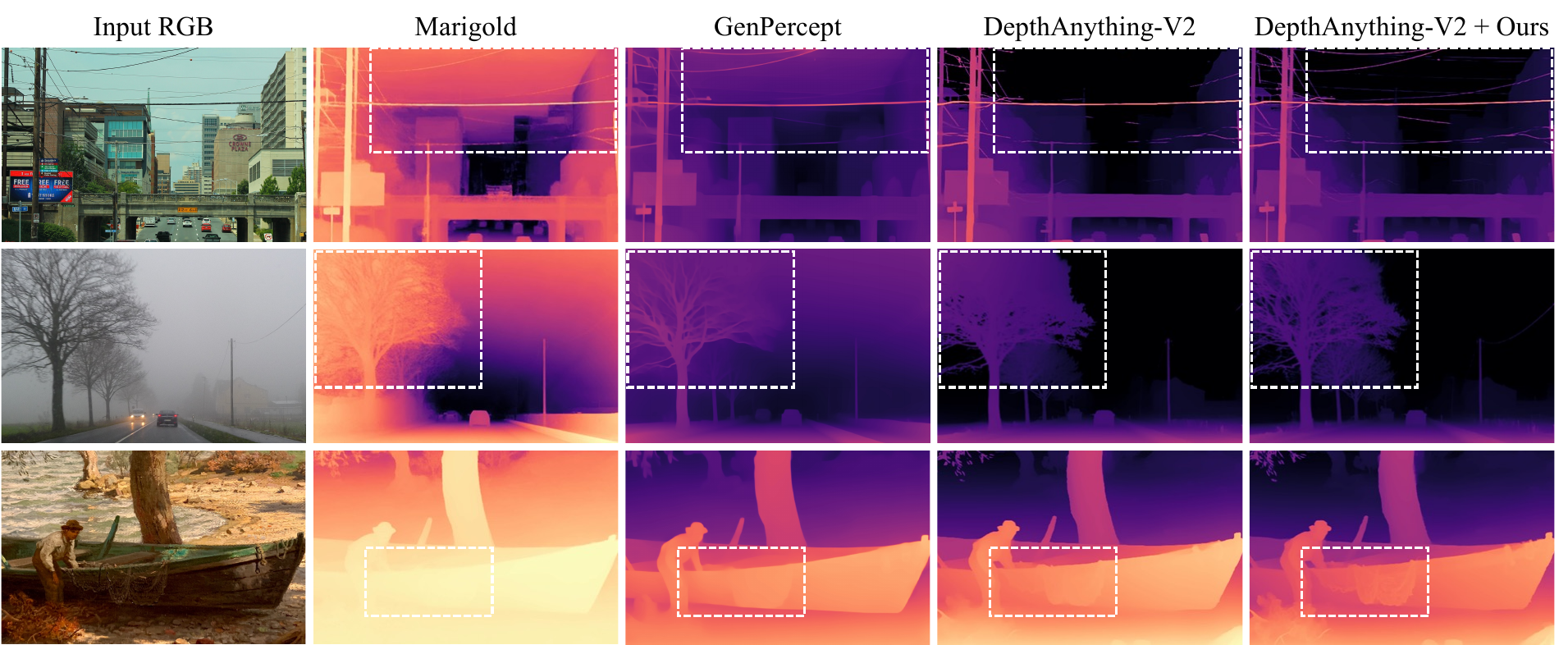}
    \vspace{-0.5em}
    \caption{Qualitative results on unseen datasets of BriGeS (Ours) and various state-of-the-art methods.}
    \label{fig:comparison}
    \vspace{-0.5em}
\end{figure*}
\begin{table*}[t]
\renewcommand{\arraystretch}{0.95}
\caption{Ablation study on proposed modules. (a) The baseline model DepthAnything-V1 \cite{yang2024depth}. (b) The baseline with Bridging Gate. (c) Our method includes the Bridging Gate and the Attention Temperature Scaling. (\textbf{Bold}: best, \underline{underlined}: second best)}
\vspace{-1.0em}
\resizebox{\textwidth}{!}{
\centering
\begin{tabular}{cccccccccccc}
\toprule
\multirow{2}{*}[-0.5ex]{Model size} & \multirow{2}{*}[-0.5ex]{\#} & \multicolumn{2}{c}{Component} & \multicolumn{2}{c}{KITTI \cite{geiger2012we}} & \multicolumn{2}{c}{NYUv2 \cite{silberman2012indoor}} & \multicolumn{2}{c}{ETH3D \cite{schops2017multi}} & \multicolumn{2}{c}{DIODE \cite{vasiljevic2019diode}}      \\
\cmidrule(lr){3-4}\cmidrule(lr){5-6}\cmidrule(lr){7-8}\cmidrule(lr){9-10}\cmidrule(lr){11-12}
           &     & Bridging Gate & Temperature Scaling  & AbsRel\textdownarrow          & $\delta$1\textuparrow         & AbsRel\textdownarrow          & $\delta$1\textuparrow         & AbsRel\textdownarrow          & $\delta$1\textuparrow         & AbsRel\textdownarrow          & $\delta$1\textuparrow         \\ \midrule
           & (a) &               &                      & 0.080                         & 0.941                         & 0.045                         & 0.979                         & 0.055                         & 0.978                         & 0.287                         & 0.754                         \\
Base       & (b) & \checkmark    &                      & \underline{0.078}             & \underline{0.944}             & \underline{0.043}             & \textbf{0.982}                & \underline{0.054}             & \underline{0.979}             & \underline{0.246}             & \textbf{0.760}                \\
           & (c) & \checkmark    & \checkmark           & \textbf{0.074}                & \textbf{0.947}                & \textbf{0.042}                & \textbf{0.982}                & \textbf{0.051}                & \textbf{0.980}                & \textbf{0.243}                & \underline{0.756}             \\ \midrule
           & (a) &               &                      & 0.080                         & 0.947                         & \underline{0.042}             & 0.981                         & 0.057                         & 0.982                         & 0.260                         & 0.758                         \\
Large      & (b) & \checkmark    &                      & \textbf{0.076}                & \textbf{0.950}                & \underline{0.042}             & \textbf{0.984}                & \underline{0.054}             & \textbf{0.983}                & \underline{0.226}             & \underline{0.760}             \\
           & (c) & \checkmark    & \checkmark           & \textbf{0.076}                & \textbf{0.950}                & \textbf{0.040}                & \textbf{0.984}                & \textbf{0.050}                & \textbf{0.983}                & \textbf{0.222}                & \textbf{0.761}                \\ \bottomrule
\end{tabular}
}
\label{tab:ablation_studies}
\vspace{-1.0em}
\end{table*}

\subsection{Comparison to State-of-the-art Methods}
We evaluate our BriGeS on zero-shot depth estimation through both quantitative and qualitative comparisons.
\tabref{tab:eval} provides a comparison of our model with six state-of-the-art relative depth estimation methods including MiDaS \cite{ranftl2020towards}, DPT \cite{ranftl2021vision}, Marigold \cite{ke2024repurposing}, GenPercept \cite{xu2024matters}, DepthAnything-V1 \cite{yang2024depth} and V2 \cite{yang2024depthv2} across multiple benchmark datasets \cite{geiger2012we, silberman2012indoor, schops2017multi, vasiljevic2019diode}.
For the DA-2K benchmark \cite{yang2024depthv2}, we extend the evaluation to include nine representative methods, covering both relative and metric depth estimation approaches.
In addition to six models above, we include UniDepth \cite{piccinelli2024unidepth}, Metric3d-V2 \cite{hu2024metric3d}, and DepthPro \cite{bochkovskii2025depth}, as detailed in \tabref{tab:eval_da2k}.

\noindent\textbf{Quantitative Comparison.}
As shown in \tabref{tab:eval}, BriGeS consistently achieves enhanced performance across the majority of datasets.
Compared to DepthAnything-V1 and V2, it achieves an average reduction of 7.33\% in the AbsRel metric.
Notably, the most significant performance improvement is observed in the DIODE dataset \cite{vasiljevic2019diode}, with an AbsRel reduction of 15.33\% in the BriGeS + DepthAnything-V1-Base setup, highlighting its remarkable effectiveness.
Similarly, in \tabref{tab:eval_da2k}, BriGeS achieves improved performance on the DA-2K benchmark.
When built upon DepthAnything-V2, it achieves the highest performance on the DA-2K, surpassing other state-of-the-art methods, demonstrating that integrating geometric and semantic information significantly enhances depth estimation accuracy.


\begin{table}[t]
  \renewcommand{\arraystretch}{0.9}
  \centering
  \caption{\textbf{Ablation study on the scaling factor $\tau$ in our Attention Temperature Scaling.} (\textbf{Bold}: best)}
  \label{tab:ablation_studies_tau}
  \vspace{-1.0em}
  \resizebox{\columnwidth}{!}{%
    \begin{tabular}{@{}cccccc@{}}
      \toprule
      Temperature Scaling Factor & 2 & 2.5 & 3 & 3.5 & 4 \\ \midrule
      Avg. Rank (\textdownarrow) & 2.3 & \textbf{1.3} & 2 & 1.5 & 2.1 \\ \bottomrule
    \end{tabular}
  }
  \vspace{-1.0em}
\end{table}

\begin{figure}[t]
  \centering
  \includegraphics[width=\columnwidth]{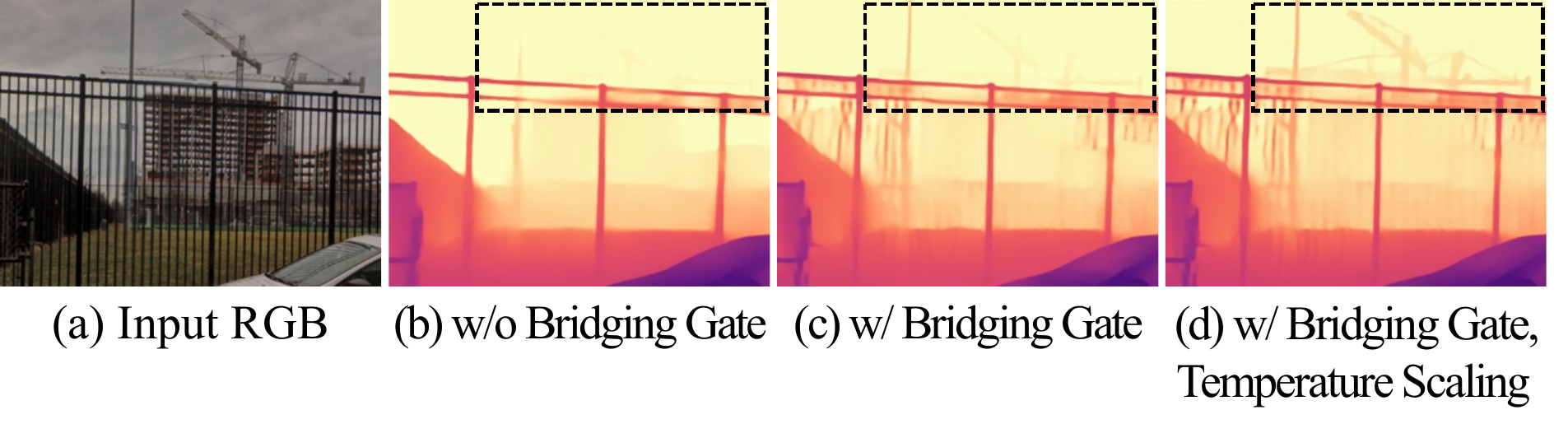}
  \vspace{-2.0em}
  \caption{Qualitative results of ablation study on proposed modules.}
  \vspace{-1.0em}
  \label{fig:ablation_v1}
\end{figure}

\noindent\textbf{Qualitative Comparison.}
\figref{fig:compare_figure_v1}, \ref{fig:comparison} show representative depth estimation results from our method.
These collectively indicate that existing models generally fail to capture semantic information for accurate depth estimation.
The first row from \figref{fig:comparison} showcases the inference results of thin power lines, which are best preserved in our method, demonstrating its capacity to recover delicate structures.
Similarly, in the second row, our model captures the intricate and thin structure of tree branches more effectively than DepthAnything-V2, while in the third row, it best preserves the fine-grained details of the fishing net.
Although Marigold and GenPercept, which leverage Stable Diffusion \cite{rombach2022high}, perform well with fine details, they mispredict the sky, as seen in the first and second rows.
These results show that our BriGeS effectively fuses geometric and semantic information, accurately delineating object edges at similar depths and preserving fine details in complex structures.

\subsection{Ablation Study}
We conduct ablation studies on BriGeS using both DepthAnything-V1-Base and Large \cite{yang2024depth} to assess the effectiveness and robustness of our proposed methods across different scales.
The quantitative results are presented in \tabref{tab:ablation_studies} for the base and large models, showcasing consistent performance improvements.
As evident in \tabref{tab:ablation_studies}-(b), our Bridging Gate effectively compensates for the lack of semantic information in the geometric features across different model sizes.
This improvement is consistently observed in both models, demonstrating the versatility and effectiveness of our Bridging Gate in enhancing semantic details irrespective of the model's scale.
Additionally, \tabref{tab:ablation_studies}-(c) demonstrates that our Attention Temperature Scaling effectively enhances the model's overall performance in both model sizes.
These results confirm that the highest performance is achieved when all proposed modules are applied, demonstrating the synergistic effect of our approaches.

\tabref{tab:ablation_studies_tau} presents the average rank calculated for the quantitative results across KITTI \cite{geiger2012we}, NYUv2 \cite{silberman2012indoor}, ETH3D \cite{schops2017multi}, and DIODE \cite{vasiljevic2019diode} by varying the temperature scaling factor for BriGeS based on DepthAnything-V1-Large.
We empirically set the scaling factor to 2.5, as it yields the best performance.
The qualitative results in \figref{fig:ablation_v1} further underscore the impact of our methods.
Notably, without the scaling module, the tower cranes within the black box were not well predicted, an issue that is effectively resolved with the scaling module.
These results demonstrate that the Bridging Gate and the scaling module help distribute attention more effectively.
By mitigating excessive focus on central objects, they reduce prediction errors associated with small structures, thereby enhancing overall performance.

\section{Conclusion}
In this work, we introduce Bridging Geometric and Semantic (BriGeS), an effective approach designed to enhance Monocular Depth Estimation (MDE) by integrating geometric and semantic information through foundation models.
The core of our method, the Bridging Gate, utilizes cross-attention and self-attention mechanisms to merge depth and semantic data effectively.
Additionally, our Attention Temperature Scaling technique ensures a balanced distribution of attention across varied inputs.
By leveraging pre-trained models, BriGeS requires minimal training data, thereby achieving both resource efficiency and strong robustness across diverse scenarios.
Our comprehensive evaluations demonstrate that BriGeS significantly outperforms existing methods, particularly in complex scenes, thus, we establish a new standard for the MDE foundation model by effectively combining geometric and semantic insights.
However, its reliance on two foundation models entails a trade-off in memory efficiency, which we aim to address by distilling knowledge into an integrated encoder that directly produces a semantic-aware geometric representation in the future.


{
\bibliographystyle{IEEEtran}
\bibliography{main}
}

\end{document}